\documentclass[10pt, conference, compsocconf]{IEEEtran}
\usepackage{booktabs}

\usepackage{cite}
\ifCLASSINFOpdf
\usepackage[pdftex]{graphicx}
\graphicspath{ {/} }
\else
\fi
\usepackage[cmex10]{amsmath}

\usepackage{caption}
\begin{document}
\title{Automatic Pixelwise Object Labeling for Aerial Imagery Using Stacked U-Nets}


\author{\IEEEauthorblockN{Andrew Khalel}
\IEEEauthorblockA{Raisa Energy LLC\\
Faculty of Engineering, Cairo University\\
Cairo, Egypt\\
andrewekhalel@eng.cu.edu.eg}
\and
\IEEEauthorblockN{Motaz El-Saban}
\IEEEauthorblockA{Raisa Energy LLC\\
Faculty of Computers and Information, Cairo University\\
Cairo, Egypt\\
melsaban@raisaenergy.com}
}


%Telephone: (800) 555--1212, Fax: (888) 555--1212}


\maketitle

\begin{abstract}
Automation of objects labeling in aerial imagery is a computer vision task with numerous practical applications. Fields like energy exploration require an automated method to process a continuous stream of imagery on a daily basis.
In this paper we propose a pipeline to tackle this problem using a stack of convolutional neural networks (U-Net architecture) arranged end-to-end. Each network works as post-processor to the previous one. Our model outperforms current state-of-the-art on two different datasets: Inria Aerial Image Labeling dataset and Massachusetts Buildings dataset each with different characteristics such as spatial resolution, object shapes and scales. Moreover, we experimentally validate computation time savings by processing sub-sampled images and later upsampling pixelwise labeling. These savings come at a negligible degradation in segmentation quality. Though the conducted experiments in this paper cover only aerial imagery, the technique presented is general and can handle other types of images.

\end{abstract}


% \fi
\IEEEpeerreviewmaketitle

\section{Introduction}
Since the introduction of high spatial and temporal resolution aerial imagery services, aerial imagery became one of the most important components in various industries. Energy, mining, civil, defense and more industries are able to use aerial imagery to enhance their productivity and quality of work. One of the most costly and time consuming tasks needed for using aerial imagery is the labeling task. In addition, hand labeling for objects of interest requires domain expertise for perfect labeling.

Recent breakthroughs in image understanding techniques using deep learning approaches together with the leap in hardware technologies specifically GPUs opened the door for more researchers to experiment with different approaches and techniques. Deep learning approaches are now deployable in industry at an affordable cost.

Generally, object labeling can be approached in two ways: localization and segmentation. Localization is where a bounding box is drawn around the detected object. The problem in localization is that the bounding box is not actually representing object's borders and can't describe its shape. On the contrary, segmentation is labeling pixels/super pixels representing the object. This will result in a very detailed detection for shape, size and outline of the object. Since objects of interest usually don't have a uniform or a fixed shape, we chose the segmentation approach for more accurate results.

This paper has two main contributions: First, we introduce a new DCNN semantic image segmentation architecture based on stacked U-nets where each network enhances the results of previous one. In our experiments on aerial imagery a cascade of two U-Nets was sufficient to outperform current state-of-the-art on two different datasets each with different characteristics. Secondly, We experiment the effect of image spatial resolution on our model performance. We find that downscaling of original resolution can decrease computation time significantly at the expense of a negligible loss in segmentation quality.

The next sections are organized as follows. Section 2 illustrates related work in the literature of semantic image segmentation. Section 3 details our methodology to approach a high quality segmentation. Section 4 presents different experiments and results to demonstrate the power of our methodology. Finally, paper conclusion and direction for future work are presented in section 5.

%-------------------------------------------------------------------------
\begin{figure*}
  \begin{center}
  \includegraphics[width=\textwidth]{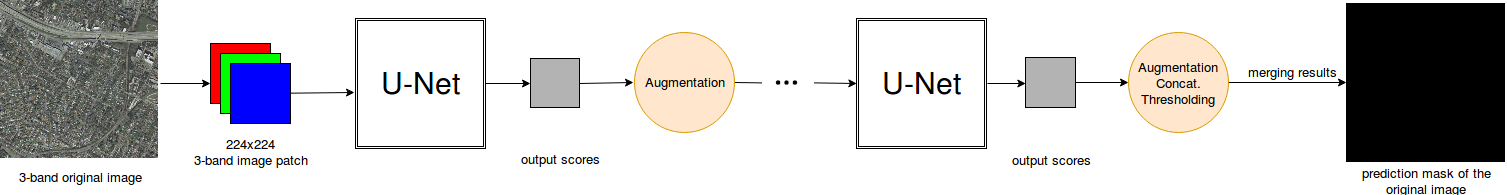}
  \caption{Full system pipeline overview}
  \label{fig:modelArch}
  \end{center}
\end{figure*}

\section{Related work}
Semantic image segmentation is the process of partitioning the image into meaningful parts, each part belongs to one of the pre-specified classes. Approaches used for semantic segmentation can be divided into: traditional and deep learning approaches. Traditional approaches usually depend on domain knowledge to extract features and apply these features to techniques like: Texton Forests \cite{shotton2008semantic}, Random Forests \cite{shotton2013real,schroff2008object}, SVM \cite{yang2012layered,felzenszwalb2010object} and Conditional Random Fields (CRFs) \cite{russell2009associative}.

One of the challenges semantic segmentation can help addressing is extracting objects of interest from the scene. The problem we are trying to address in this paper is to extract buildings from aerial imagery which has been attempted previously using different approaches. Many of these approaches use hand-crafted features, classifiers and boosting \cite{inglada2007automatic,porway2008hierarchical,senaras2013building} or contour detection to find a rectangular (building-like) objects \cite{kim1999development}.

Li et al. \cite{li2015robust} uses the unsupervised Gaussian mixture model (GMM) to segment the image into homogeneous super pixels and then Higher order Conditional Random Field (HCRF) is used for accurate rooftop extraction. Jin and Davis \cite{jin2005automated} generate building hypothesis using edge-based segmentation methods and verified them using differential morphological profile (DMP). Buildings are extracted using structural, contextual, and spectral information.

In recent years deep learning methods have shown excellent performance in many fields including semantic segmentation. Long et al. \cite{long2015fully} introduced the Fully Connected Networks (FCN) as an end-to-end architecture producing dense output maps. Long also introduced the concept of upsampling using deconvolutional layers. Since object location is very important in semantic segmentation - unlike in classification - new architectures were developed to preserve this location information.

First set of architectures are inspired by the idea of encoder-decoder where the input image is encoded into smaller intermediate form using pooling layers and then recovered to original size using upsampling layers in the decoder usually with help of skip connections from encoder to decoder. Most popular architectures of this set are: (1) U-Net introduced by Ronneberger et al. \cite{ronneberger2015u} originally for medical image segmentation. U-Net won the ISBI cell tracking challenge 2015 with a large margin. (2) SegNet \cite{badrinarayanan2015segnet} which doesn't use skip connections and saves the pooling indices to be used in the decoder for non-linear upsampling.

A different set of architectures depends on atrous (also called dilated) convolution \cite{chen2016deeplab} instead of pooling layers. In atrous convolution filters are "with holes" so that we can enlarge the receptive field of the filter without decreasing the image spatial resolution.

Volodymyr Mnih uses convolutional neural networks in his PhD thesis \cite{MnihThesis} to train an aerial image labeling system for roads and buildings. He tried neural networks and Conditional Random Fields as post-processing to CNN. His model shows good performance on Massachusetts roads and buildings datasets \cite{MnihThesis}.

Saito and Aoki \cite{saito2015building} use CNN for road and building detection. They use the normal downsampling architecture of the CNN and at the end, a fully connected layer with Dropout \cite{srivastava2014dropout} is added to infer prediction of the input image. Their model outperforms Mnih's models \cite{MnihThesis} for both roads and buildings using a single model for each class.

Newell et al. \cite{newell2016stacked} propose a network architecture that reaches state-of-the-art results in human pose estimation. They call their architecture Hourglass due to its shape of contracting and expanding paths. This architecture is very similar to U-Net, it only differs in the way tensors are concatenated. Hourglass uses addition operator to add the two tensors together into a new sum tensor. Human pose estimation problem can be formulated as the task of extraction joints from an input image. With this formulation, we can adapt the same network architecture for the task of semantic image segmentation.

\section{Methodology} \label{section:methodology}
An overview of our full pipeline is shown in Figure \ref{fig:modelArch}. Our pipeline starts by dividing the input image into smaller patches 224x224x3 pixels. These patches are the input of our model and the output is a cropped prediction mask. By concatenating these small outputs, we can get a full size prediction mask. More levels of U-Nets are used to enhance the results. %This setup primarily works on enhancing the edges of the detected buildings and filling holes within the same building.
\subsection{Network architecture}
\begin{figure*}
  \begin{center}
  \includegraphics[width=\textwidth]{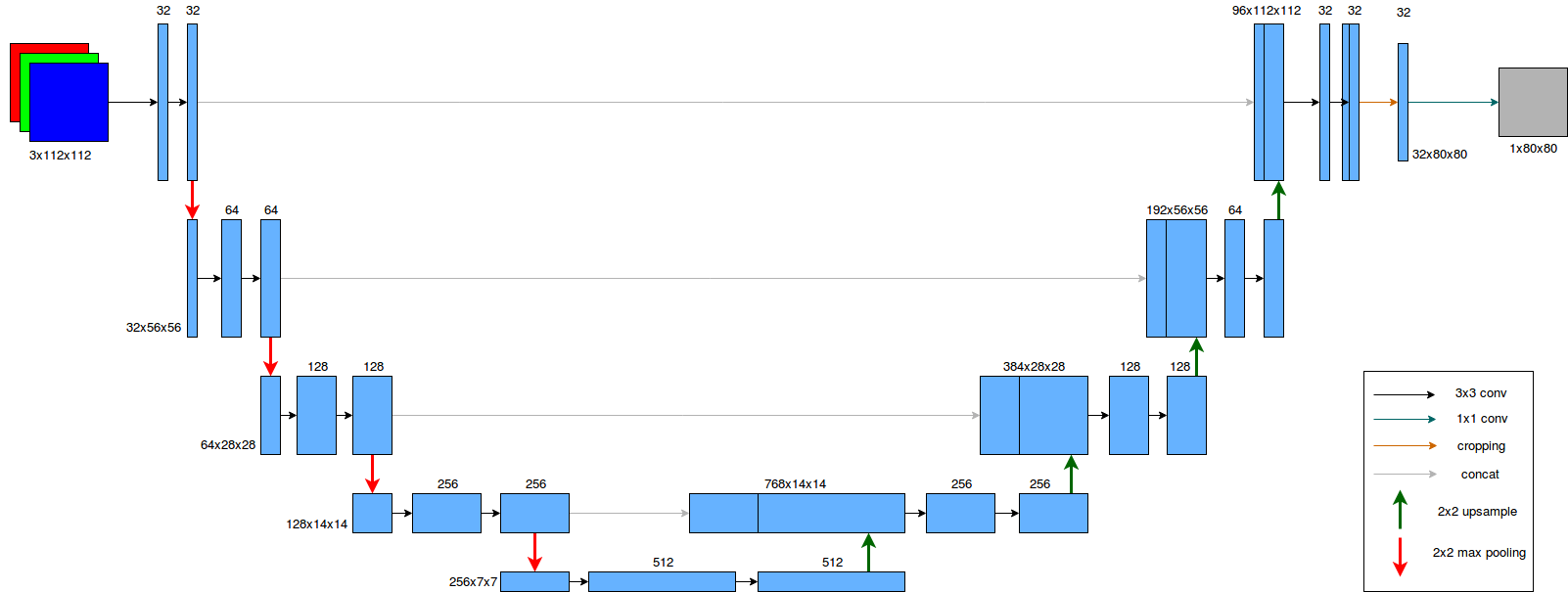}
  \caption{Detailed architecture of our U-Net}
  \label{fig:network}
  \end{center}
\end{figure*}
As shown in Figure \ref{fig:network} most of our layers consist of 3x3 convolution filters since they are computationally efficient. Filters' count double as we go deeper in the contracting path while they are halved while going through the expanding path. Each layer also has Batch Normalization \cite{ioffe2015batch} layer for faster convergence. Max pooling with size of 2x2 is used for down sampling while for up-sampling, elements in the original tensor are replicated to its 2x2 window in the output up-sampled tensor. Concatenation is done by appending the two tensors into a new activation volume. Finally, He uniform variance scaling initializer \cite{he2015delving} is used for all convolutional layers.

The partitioning of the whole image into smaller patches will cause buildings on the patch edges to lose important parts of their structure which leads to poor performance at edges. This problem can be solved in two ways: using overlapped patches or using cropping layer in our network. We used cropping layer as it turns out to be more effective solution \cite{iglovikov2017satellite}.

\subsubsection{Training}
Nadam optimizer \cite{dozat2016incorporating} is used to train the model. For the first level U-Net a learning rate of 1e-3 is used for 50 epochs and then 1e-4 is used for another 50 epochs.  A batch size of 128 patches is used. The second level U-Net uses a learning rate of 1e-4 and is trained for 50 epochs. Since Intersection over Union (IoU) becomes the standard metric in semantic image segmentation \cite{long2015fully} and it is non-differentiable, a joint loss function $L$ proposed by Iglovikov et al. \cite{iglovikov2017satellite} is used to combine both a differentiable form of IoU and binary cross entropy
$$
H = -\frac{1}{n}\sum_{i=1}^{n}[y\log(\hat{y}) + (1-y) \log(1-\hat{y})]$$
$$J(y,\hat{y}) = \frac{1}{n}\sum_{i=1}^{n}\frac{y_{i}\cdot\hat{y_{i}}}{y_{i}+\hat{y_{i} - y_{i}\cdot\hat{y_{i}}}}
$$
$$
L = H - \log(J)
$$
where $n$ is the number of images in a batch, $y$ is the ground-truth value and $\hat{y}$ is the prediction.

Moreover, data augmentation is applied at training time by choosing randomly from a set of transformations: horizontal flip, vertical flip and rotations. Data augmentation helps in building a strong model which is less dependent on input image orientation. This is very helpful for our model to generalize to different regions other than regions in training set.

\subsubsection{Prediction}
To make more confident predictions, test time augmentations are applied where the same set of transformation applied at training time is applied to each image patch before prediction. The predictions of all transformed versions are averaged. This average is the final prediction score. Then, thresholding is applied to convert scores into binary values of the mask. The threshold value is a hyperparameter which we tuned using cross validation set. %The best value for the threshold was 0.5.

To reduce discontinuity effect of image sub-divsion into tiles we use use image mirroring as proposed by Ronneberger et al. \cite{ronneberger2015u}. This yields better results at tile boundaries.
% one of the problems that arise with using image patches is behavior at image tile boundaries. Ronneberger et al. \cite{ronneberger2015u} addressed this problem by using an edge mirroring technique to prevent discontinuity at boundaries. We used the same technique for each tile and it yields better results at the boundaries.
Our pipeline is built using Keras \cite{chollet2015keras} library with Theano \cite{2016arXiv160502688short} as backend.
\section{Results}

In this section we introduce the datasets used in experimentation and report on conducted experiments and their results.

\begin{figure*}
  \begin{center}
  \includegraphics[width=\textwidth]{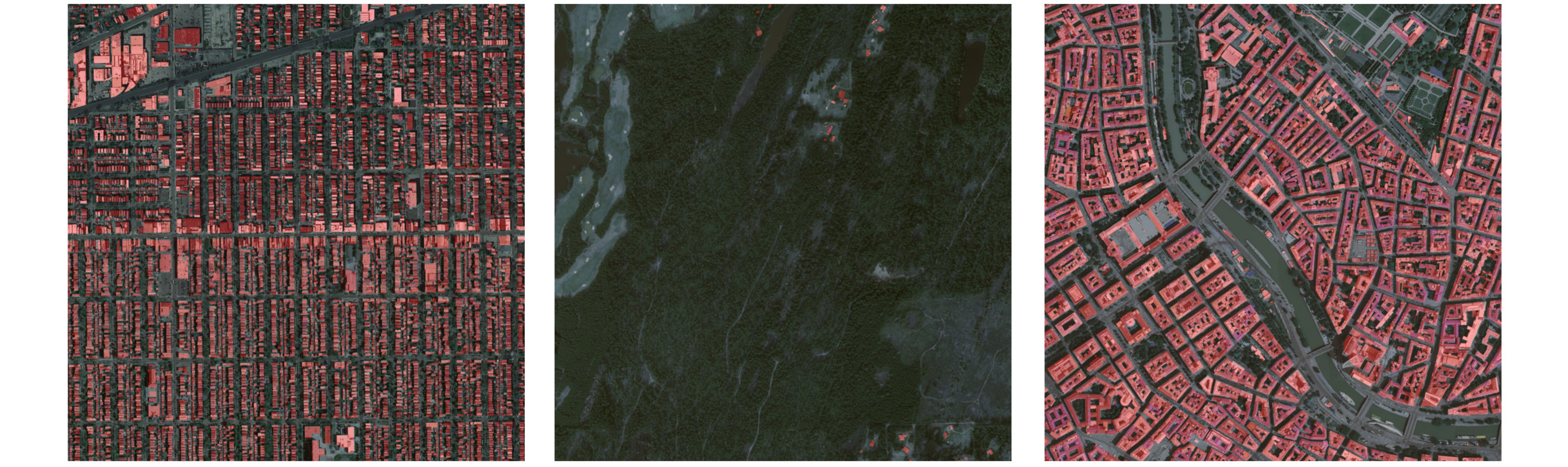}
  \caption{Different urban densities in the dataset. Chicago, Kitsap County, and Vienna respectively. Red areas represent buildings' ground truth labels}
    \label{fig:urbanDensities}
  \end{center}
\end{figure*}

\subsection{Datasets}
To illustrate the power of the proposed model we use two datasets: Inria Aerial Image Labeling dataset \cite{maggiori2017dataset} and Massachusetts Buildings dataset \cite{MnihThesis}. We selected these two datasets because they cover different imagery characteristics such as spatial resolution, object types, shapes and sizes.

Inria's dataset is specifically constructed to address automatic pixelwise labeling of aerial imagery. The dataset consists of two subsets: training and testing sets. Each subset covers 405 km$^2$ area with spatial resolution of 0.3 m. The provided data are 3-band colored orthorectified images. Training data is labeled for two classes: building and not building. Training dataset covers Austin, Chicago, Kitsap County, Western Tyrol and Vienna, while test set covers a set of different regions: Bellingham, Bloomington, Innsbruck, San Francisco, Eastern Tyrol. For each region in the two subsets, there are 36 tiles of size 5000x5000 pixels that cover 1500x1500 m area. A sample from the dataset images and labels is shown in Figure \ref{fig:datasetSample}.
\begin{center}
  \includegraphics[width=0.45\textwidth]{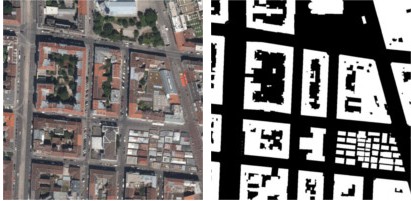}
  \captionof{figure}{A sample of Inria's dataset from authors' website
   \cite{datasetWebsite} aerial images (on the left) and buildings' ground truth mask (on the right)}
  \label{fig:datasetSample}
\end{center}
There are two important aspects to choose Inria's dataset for our experiments. First, training and testing sets cover different regions so we will be able to judge the ability of our model to generalize to new regions. Secondly, the covered regions are very different in their urban densities. Figure \ref{fig:urbanDensities} shows the large variety of urban densities in different regions. Chicago has very dense and small buildings. Kitsap county has a very sparse distribution of buildings due to its large green areas. Vienna has a very different architecture style: large buildings without a full roof. This variability in the dataset ensures that the model will learn to label different regions and understands the structure of a building in a more general sense.

The second dataset we use is the Massachusetts buildings dataset. It consists of 151 aerial images of size 1500x1500 pixels covering urban and suburban regions at the area of Boston. Each image covers an area of 2.25 km$^2$ at a resolution 1 m$^2$/pixel. These images were randomly split into training, validation and test sets with sizes 137, 4 and 10 respectively. A sample of the dataset is shown at Figure \ref{fig:mnihSample}.

\begin{center}
  \includegraphics[width=0.47\textwidth]{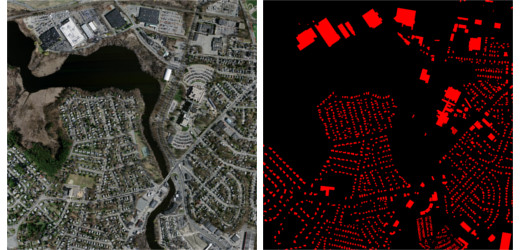}
  \captionof{figure}{A sample of Massachusetts buildings dataset with aerial image and buildings' mask.}
  \label{fig:mnihSample}
\end{center}

In order to enable comparison with results from other researchers, we use the same performance measures of each dataset. Inria's dataset uses two main performance measures which are: Intersection over Union (IoU) and Accuracy. Intersection over union, also known as Jaccard index is defined as:
$$ IoU(GT,P) = \frac{\text{Area of Intersection between GT and P}}{\text{Area of Union between GT and P}} $$
where $GT$ is the ground truth mask and $P$ is the predicted mask. Accuracy is defined as:
$$ Accuracy(GT,P) = \frac{\text{Area of correctly classified pixels}}{\text{Area of  GT}}$$
We focused our experiments on IoU as it becomes a standard for semantic segmentation \cite{long2015fully}. Moreover, accuracy is not discriminative enough since large image areas are dedicated to background (non-building) class.

For Massachusetts buildings dataset, a relaxed version of precision and recall is used to calculate the precision-recall breakeven point \cite{MnihThesis}. The relaxation assumption is to consider a positive label correct if it falls within the 7x7 region of any ground truth positive pixel. Since the buildings' masks are usually not perfectly aligned to the image, this relaxation will provide a realistic performance measure.

% \subsubsection{Perfect labels?}
% \begin{center}
%   \includegraphics[width=0.4\textwidth]{noisy1}
%   \captionof{figure}{Labeling mistakes in chicago2.tif}
%   \label{fig:noisy1}
% \end{center}
% We started with an exploration for the training images and labels. Visualizing images overlayed with label masks shows that the labels aren't perfect. In Figure \ref{fig:noisy1} we can see the building labels are shifted left from the where they should be. Moreover, on the right we can see two cars are labeled as a building. Figure \ref{fig:noisy2} shows a lot of building labels for a an area with no buildings at all.
% \begin{center}
%   \includegraphics[width=0.4\textwidth]{noisy2}
%   \captionof{figure}{False building labels in kitsap4.tif}
%   \label{fig:noisy2}
% \end{center}
% Although the dataset labels aren't perfect but these mistakes are tolerable due to the large volume of training images. So mistakes wouldn't have a tremendous effect on results.
\subsection{Best model results}
After running through all experiments and choosing the best model described in section \ref{section:methodology}, We compare our results with the results of other approaches as shown in table \ref{table:testResults}.
% We run the best model on the test set of Inria's test set which consists of 180 images and submit the results to the contest. Our model scored the highest IoU of value 65.94 and a margin of 4.21 to the nearest competitor on the \href{https://project.inria.fr/aerialimagelabeling/leaderboard/}{leaderboard}. Detailed results for each of the five regions are shown in Table \ref{table:testResults}.
\begin{table*}
\centering
 \begin{tabular}{||*8c||}
 \hline
 Method && Austin & Chicago & Kitsap Co. & West Tyrol & Vienna & Overall\\ [0.5ex]
\hline\hline
FCN + MLP \cite{maggiori2017dataset}& IoU & 61.20&61.30&51.50&57.95&72.13&64.67\\
(Baseline) & Acc. & 94.20&90.43&98.92&96.66&91.87&94.42\\
\hline
SegNet (Single-Loss) \cite{bischke2017multi} & IoU & 76.49 & 66.77 & 72.69 & 66.35 & 76.25 & 72.57\\
& Acc. & 93.12 & 99.24 & 97.79 & 91.58 & 96.55 & 95.66\\
\hline
SegNet + MultiTask-Loss \cite{bischke2017multi}& IoU & 76.76&67.06&\textbf{73.30}&66.91&76.68&73.00\\
(Uncertainty Weighted) & Acc. & 93.21&\textbf{99.25}
&97.84&91.71&\textbf{96.61}&95.73\\
\hline
% Single U-Net + aug.& IoU & 77.19 & 67.90 & 72.01 & 75.77 & 79.00 &74.37 \\
% (initial model) & Acc. & 96.70 & 92.40 & 99.24 & 98.17 & 93.97 & 96.10\\
% \hline
2-levels U-Nets + aug. & IoU & \textbf{77.29} & \textbf{68.52} & 72.84 & \textbf{75.38} & \textbf{78.72} & \textbf{74.55}\\
(Our model) & Acc.& \textbf{96.69} &92.40&\textbf{99.25}&\textbf{98.11}&93.79& \textbf{96.05}\\
%  Inria1 & 52.91 	& 	46.08 	& 	58.12 	& 	57.84 	&	59.03 	& 	55.82\\
%  \hline
%   Inria2 & 56.11 & 50.40 & 61.03 & 61.38 & 62.51 & 59.31\\
%  \hline
%  TeraDeep & 58.08 	& 	53.38 	& 	59.47 	& 	64.34 	& 	62.00 	& 	60.95\\
%  \hline
%  RMIT& 	57.30 	&	51.78 	& 	60.70 	&	66.71 	& 	59.73 	&61.73\\
%  \hline
% Our model & \textbf{64.46} 	&	\textbf{56.63} 	& 	\textbf{66.99} 	&	\textbf{67.74} 	&	\textbf{69.21} 	&	\textbf{65.94}\\
\hline
\end{tabular}
\captionof{table}{Results of different methods for Inria Aerial Image Labeling validation set}
\label{table:testResults}
\end{table*}
Figure \ref{fig:result1} shows the resulting labels of our model on an image of Innsbruck from the test set. The figure shows detections of buildings with different shapes (rectangular and non-rectangular) and sizes.
\begin{center}
  \includegraphics[width=0.45\textwidth]{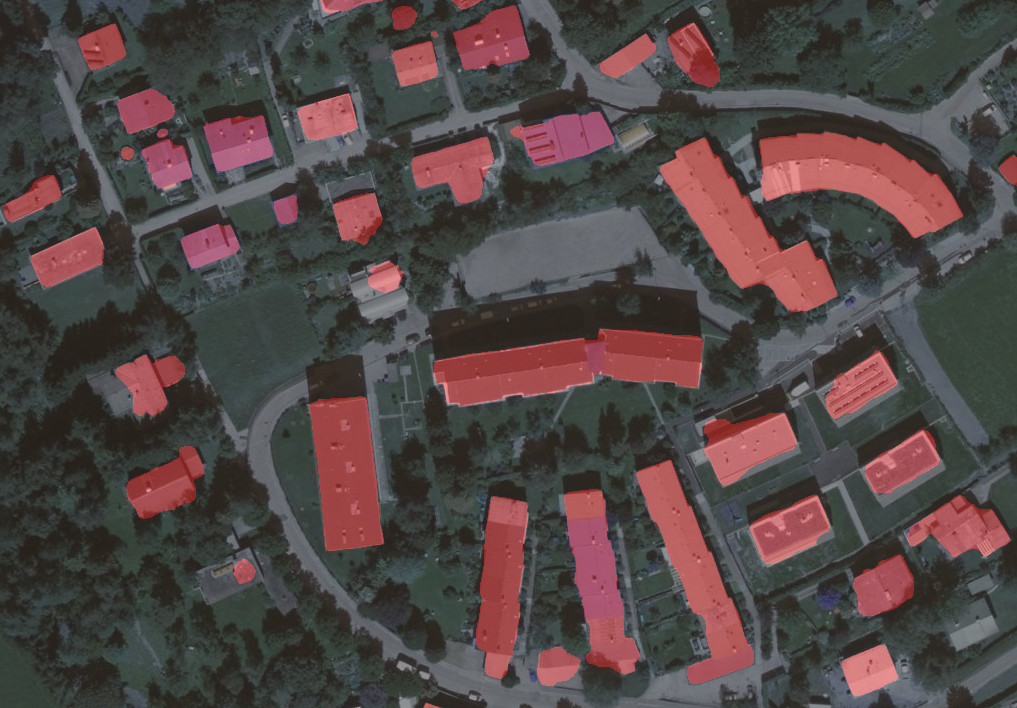}
  \captionof{figure}{Our model labels (red) for buildings with different shapes at Innsbruck}
    \label{fig:result1}
\end{center}

In some cases our model computes a wrong segmentation result. For example in figure \ref{fig:result3} we can see the model detecting a parking lot as a building due to its color which is very similar to buildings colors in this area and due to its texture which looks like house's roof.
\begin{center}
  \includegraphics[width=0.45\textwidth]{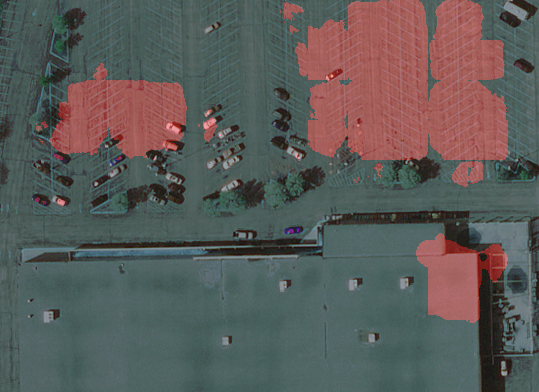}
  \captionof{figure}{False positives at Bloomington}
    \label{fig:result3}
\end{center}

Table  \ref{table:mnihResults} shows our model results on the  Massachusetts building dataset. We significantly outperform other approaches from the literature. Figure \ref{fig:mnihResults} shows a sample of our model predictions. Although the two datasets' characteristics are quite different our model has leading results on both of them with the same architecture.

\begin{table}[!htbp]
\centering
\begin{tabular}{*2c}
\toprule
Method &  Precision-recall breakevent point \\
\midrule
Mnih et al. \cite{MnihThesis} & 0.9211\\
Saito et al. \cite{saito2015building} &  0.9230\\
Marcu et al. \cite{marcu2016local} & 0.9423\\
Our model & \textbf{0.9633}\\
\bottomrule
\end{tabular}
\caption{Precision-recall breakeven point of different approaches for the Massachusetts dataset.}
\label{table:mnihResults}
\end{table}

\begin{center}
  \includegraphics[width=0.45\textwidth]{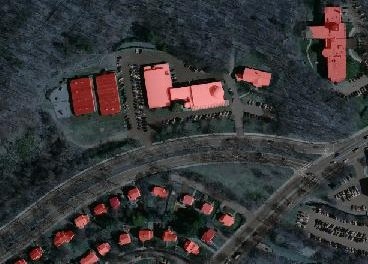}
  \captionof{figure}{Our model labels (red) for buildings on Massachusetts dataset}
    \label{fig:mnihResults}
\end{center}

The training time of the first U-Net was 41 hours on Nvidia Tesla K80 GPU, while for the second U-Net, training time  was 20.8 hours. Prediction of a single tile of size (5000x500 pixels) took 3.48 minutes including the augmentation, concatenating small patches and thresholding.

\subsection{Different architectures and pipelines}
After presenting our model best results, we present in this section several variations we have experimented with. The goal is to highlight key factors leading to the superior performance of our model. Table \ref{table:results} shows results for different conducted experiments on Inria's dataset. Using a single U-Net architecture to directly predict the image is a good starting point. Since data augmentation is very important to produce a robust model which is invariant to different rotations and orientations, we test its effect on the results by running the same model of a single U-Net with the same configurations in addition to the augmentation on both training and testing time. This change leads to an enhancement in IoU of the validation set as shown in the results.

\begin{table}[!htbp]
\centering
\begin{tabular}{*2c}
\toprule
Method & IoU \\
\midrule
Single U-Net & 73.68 \\ %70.69
Single U-Net + aug. & 74.38\\ %71.23
Single U-Net + aug. + CRF & 72.58\\ %69.64
Single Hourglass + aug. & 72.30\\ %69.37
2-level U-Nets + aug. & \textbf{74.60}\\ %71.71
\bottomrule
\end{tabular}
\caption{IoU of different pipelines for Inria Aerial Image Labeling dataset.}
\label{table:results}
\end{table}

Applying Hourglass \cite{newell2016stacked} architecture with data augmentation to our problem gives an IoU score of 72.30. Although Hourglass didn't produce better results, it guided us to the idea of networks stacking (our final pipeline). Stacked Hourglass architecture consists of multiple consecutive Hourglasses arranged end-to-end.

\subsection{Downsampling}
As mentioned before the spatial resolution of Inria's dataset is 0.3 m. We want to investigate the effect of lower resolution on the results. To try different resolutions of the same data, we re-sample the data at lower rates. Our experiments are conducted at two resolutions: $\frac{1}{2}$ and $\frac{1}{4}$ the original resolution. To ensure fair comparison between different resolutions, we upsample the lower resolution masks to the original resolution after the prediction using simple linear interpolation before calculating Jaccard index. The result of our single U-Net model with only 50 epochs of training on different resolutions is shown in Table \ref{table:diffRes} along with the prediction time per image.
\begin{table}[!htbp]
\centering
\begin{tabular}{*3c}
\toprule
Resolution &  IoU & Prediction time\\&&(secs/tile)\\
\midrule
1   & 71.23 & $\sim$160\\
1/2    & 71.07 & $\sim$40\\ %before interpolation  0.7147
1/4    &  70.71 & $\sim$17\\ %before interpolation 0.7155
\bottomrule
\end{tabular}
\caption{IoU and prediction time per full image (tile) at different resolutions using a single U-Net architecture.}
\label{table:diffRes}
\end{table}
As the table illustrates, the results for lower resolutions are very close to the results of the original resolution. However, there are substantial savings in prediction time. Also, the overhead of downsampling and upsampling is negligible ($\sim$0.06 secs/image). These findings show that a very high resolution can be replaced by a lower one for considerable gains in computation time.

% \subsection{Data augmentation effect}
% Data augmentation is very important to produce a robust model which is invariant to different rotations and orientations. To experience the effect of data augmentation on the results we ran the same model of a single U-Net with the same configurations in addition to the augmentation on both training and testing time. This change leads to an enhancement in IoU of the validation set from 70.69 to 71.23.

\subsection{Using Conditional Random Fields (CRFs) \cite{lafferty2001conditional} as post-processing}
Fully connected CRFs proved to be very effective for the localization challenge \cite{chen2016deeplab,krahenbuhl2011efficient}. They are capable of finding fine-grained edges and outlines which enhances object segmentation quality.

\begin{center}
  \includegraphics[width=0.47\textwidth]{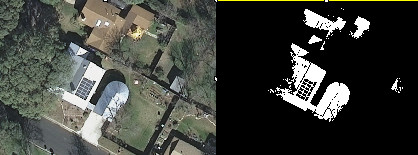}
  \captionof{figure}{detected building after CRF processing. The real satellite image is on the left while the detected mask is on the right}
  \label{fig:crf}
\end{center}

Using CRF with a single U-Net scored 72.58 on the validation set. Figure \ref{fig:crf} shows clearly how CRF draws detailed outer and inner edges of the detected building however this isn't required for our datasets. Our datasets requires a solid polygon covering the whole building without any roof details and lines. Based on these results we have not incorporated CRFs as post-processor to our pipeline. %So running CRFs after the two levels U-Nets scores 71.70 on the validation set. The score is slightly lower than the two U-Nets alone.

\section{Conclusion and future work}
In this paper we propose a stack of deep convolutional neural networks built on the U-Net \cite{ronneberger2015u} architecture to perform pixelwise labeling of aerial images. Our approach outperforms all other models on both Inria's aerial image labeling dataset \cite{maggiori2017dataset} and Massachusetts Buildings dataset \cite{MnihThesis}. In addition, experiments show that we can achieve sizable gains in processing time by working on lower resolution images. This could be very helpful for interactive applications that require fast labeling.

For future work, we will investigate suitable methods that can adapt learned models at one specific spatial resolution to work on different resolutions with minimal changes as this can be very useful for models to learn on a dataset starting from a model trained on a different dataset. Another possible future direction is to leverage Generative Adversarial Networks (GANs) \cite{goodfellow2014generative} to improve model segmentation quality through a generator-discriminator network pair.

\bibliographystyle{IEEEtran}
\bibliography{IEEEabrv,IEEEexample}

\end{document}